# Phase-based Minimalist Parsing and complexity in non-local dependencies

**Cristiano Chesi**
NETS - IUSS
P.zza Vittoria 15
I-27100 Pavia (Italy)
cristiano.chesi@iusspavia.it

## Abstract

**English.** A cognitively plausible parsing algorithm should perform like the human parser in critical contexts. Here I propose an adaptation of Earley's parsing algorithm, suitable for Phase-based Minimalist Grammars (PMG, Chesi 2012), that is able to predict complexity effects in performance. Focusing on self-paced reading experiments of object clefts sentences (Warren & Gibson 2005) I will associate to parsing a complexity metric based on cued features to be retrieved at the verb segment (Feature Retrieval & Encoding Cost, FREC). FREC is crucially based on the usage of memory predicted by the discussed parsing algorithm and it correctly fits with the reading time revealed.

**Italian.** *Un algoritmo di parsing cognitivamente plausibile dovrebbe avere una performance paragonabile a quella umana in contesti critici. In questo lavoro propongo un adattamento dell'algoritmo di Earley che utilizza Grammatiche Minimaliste basate sul concetto di Fase (PMG, Chesi 2012). Associata all'algoritmo, verrà discussa una funzione di costo (Feature Retrieval & Encoding Cost, FREC) capace di misurare la difficoltà relativa al recupero dei referenti coinvolti in dipendenze a distanza. La funzione si basa sui tratti morfosintattici archiviati nel memory buffer utilizzato dal parser. Concentrandosi sulle strutture scisse ad estrazione dell'oggetto, si mostrerà come il FREC risulti predittivo dei dati sperimentali ricavati da studi classici di lettura autoregolata (Warren & Gibson 2005).*

## 1 Introduction

The last twenty years of formal linguistic research have been deeply influenced by Chomsky's minimalist intuitions (Chomsky 1995, 2013). In a nutshell, the core Minimalist proposal is to reduce phrase structure formation to the recursive application of a binary, bottom-up, structure-building operation dubbed *Merge*. Merge creates hierarchical structures by combining two lexical items (1.a), one lexical item and an already built (by previous application of Merge operations) phrase (1.b) or two already built phrases (1.c).

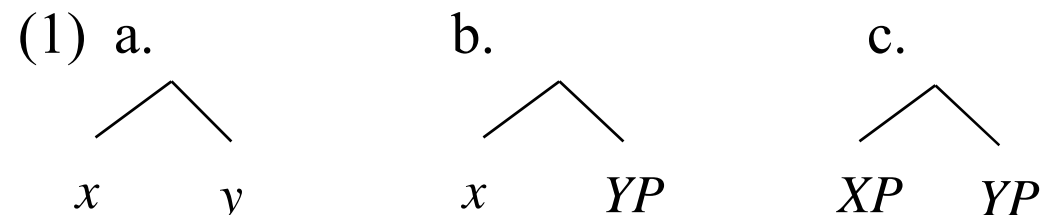

Phrases are not linearly ordered by Merge. Only when they are spelled-out (i.e. sent to the Sensory-Motor interface, aka Phonetic Form, *PF*), linearization is required: assuming that *x* and *y* are terminal nodes (i.e. words), either <*x*, *y*> or <*y*, *x*> can both be proper linearizations of (1.a). Hierarchical structure (and linearization) is also determined by another structure building operation: *Move* (or *Internal Merge*, Chomsky 1995); Move re-arranges phrases in the structure by re-merging an item (already merged in the structure) to the edge of the current, top-most, phrase: for instance [*XP* [*YP* [*ZP*]]] can lead to [*ZP* [*XP* [*YP* (*ZP*)]] if *XP* (the *probe*) has a feature triggering movement (e.g. +*f*) and *ZP* (the *goal*) has the relevant feature qualifying it as a plausible target for movement (e.g. -*f*). At the end, the element displaced (*ZP*) will occupy the edge of the structure. When the items within an already built phrase, for instance *XP*, are delivered to PF, they get properly linearized according to their hierarchical structure (e.g. Linear Correspondence Axiom, Kayne 1994), intrinsic phonetic properties (e.g. cliticization), as well as economy conditions (e.g. an items should not be pronounced twice). Such a (cyclic) spell-out happens at *phases*: *XP* will be delivered to PF only if it qualifies as a phase (Chomsky 2013). In this sense, a phase should be a constituent/phrase with some degree of completeness with respect to semantic interpretation (Logic Form, aka LF). Most minimalist linguists agree on the fact that a full-fledged sentence (aka Complementizer Phrase, *CP*) is a phase, the highest argumental shell of a predicate qualifies as a phase (aka little-*v* Phrase, *vP*) and also a full argument is a phase (aka Determiner Phrase, *DP*). Such a simple (and computationally appealing) model has been fully formalized (Stabler 1997, Collins & Stabler 2016) and some parsing algorithm that implements main minimalist insights has been discussed in literature (e.g. Harkema 2001, Chesi 2012 a.o.).

In these pages, I will present some of the advantages of retaining such a simplified computational approach to syntactic derivation. Crucially, I will try to overcome some clear disadvantages in assuming the just presented standard, bottom-up, structure building operations, while obtaining, at the same time, a better empirical fit: on the one hand, I will avoid any non-efficient deductive-parsing perspective (that is a consequence of the assumed bottom-up nature of the Merge and Move operations); on the other, I will promote a more transparent relation between formal competence, parsing and psycholinguistic performance by presenting a simple adaptation of Earley's Top-Down parsing algorithm (Earley 1970) and a complexity metric that refers directly to parsing memory usage: this metric will be able to account for complexity in retrieving the correct item while processing specific non-local dependencies. By "non-local" dependencies I refer to those relations involving movement, namely constructions where the very same item occurs in two distinct, non-adjacent, positions: for instance, *wh*-dependencies in English require the *wh-* item (*who*, in (1)) to be interpreted both in a the left peripheral (focalized) position (the Criterial position, in the sense of Rizzi 2007) and in the thematic lower position (right next to the verb *meet* in (1))[1]:

(1) *Who*$_1$ do you think Mary will meet _$_1$?

The critical derivation I will discuss in this paper is that of object clefts (Gordon et al. 2001) that with *wh*-questions share a similar non-local dependency formation:

(2) a. It is [$_{DP1}$ the banker|John|me] that [$_{DP2}$ the lawyer|Dan|you] will meet _$_{DP1}$

In short, the head of the dependency (DP1) should be interpreted both as a focalized item and as the direct object (this is where the name of the construction "object cleft" comes from) of the embedded verb. The difficulty of parsing this structure has been deeply discussed in literature (Gordon et al. 2004). What is considered a crucial factor is the role of the similarity between $DP_1$ and $DP_2$ (the subject of the cleft, Belletti and Rizzi 2013, §2). To capture this fact, I will re-adapt Earley's algorithm (§3.1) to operate on a specific version of Minimalist Grammar (§3). This would allow us to subsume the similarity effect by predicting reading differences as revealed in self-paced reading experiments (e.g. Warren & Gibson 2005, §4).

## 2 Parsing with Minimalist Grammars

Since Merge and Move strictly operate "from bottom to top", we expect sentence structure in (2) to be built in 9 steps (and 5 phases: *ph1*, *ph2* …):

1. [$_{ph1}$ the banker]
2. [$_{ph3}$ meet [$_{ph1}$ …]]]
3. [$_{ph3}$ will [meet [$_{ph1}$ …]]]
4. [$_{ph2}$ the lawyer] (independently built)
5. [$_{ph3}$ [$_{ph2}$ …] will [meet [$_{ph1}$ …]]]
6. [$_{ph4}$ that [$_{ph3}$ [$_{ph2}$ …] will [meet [$_{ph1}$ …]]]]
7. [[$_{ph1}$ …] [$_{ph4}$ that [$_{ph3}$ [$_{ph2}$ …] will [meet ($_{ph1}$ …)]]]]
   (*ph*$_1$ moves to *ph*$_4$ edge)
8. [$_{ph5}$ is [[$_{ph1}$ …] [$_{ph4}$ that [$_{ph3}$ [$_{ph2}$ …] will [meet [$_{ph1}$ …]]]]]
9. [$_{ph5}$ it [is [[$_{ph1}$ …] [$_{ph4}$ that [$_{ph3}$ [$_{ph2}$ …] will [meet [$_{ph1}$ …]]]]]

With the exception of step 4, all other steps must be strictly ordered. As a consequence, moving the direct object in the relevant position would force the linearization to place $ph_2$ first at the edge of $ph_3$, then at the edge of $ph_4$. This is how Minimalism derives the relevant non-local dependencies in (2). Obviously this is not transparent at all with respect to parsing (e.g. Fong 2011), where the processing order is expected to be completely reversed:

1. [$_{ph5}$ ] is initiated
2. [$_{ph1}$ ] is fully processed while [$_{ph5}$ ] is still open

[1] Coreference in non-local dependencies will be indicated by the same subscript placed both on the "displaced" item and on the thematic position (the non-pronounced item in the thematic position is indicated with a co-indexed underscore)

3. [$_{ph4}$ ] is initiated (a Relative Clause)
4. [$_{ph3}$ ] is initiated as well (Verbal Phrase)
5. [$_{ph2}$ …] is fully processed while [$_{ph5}$ ], [$_{ph4}$ ] and [$_{ph3}$ ] are open
6. [$_{ph1}$ ] finally receives a thematic role, hence [$_{ph5}$ ], [$_{ph4}$ ] and [$_{ph3}$ ] can be closed.

Unless we deeply revise Minimalist Grammars (both with respect to movement, Fong 2005, and to thematic role assignment, Niyogi & Berwick 2005), we are left with an asymmetry that can not be explained simply in terms of structure building operations as discussed in the next section.

### 2.1 The "similarity" problem

Warren & Gibson (2005) show that in clefts constructions like the one discussed in (2), the variation of the two DPs [$_{ph1}$ ] and [$_{ph2}$ ] produces differences in reading time at the verb segment in self-paced reading experiments with the full-DP matching condition ([$_{ph1}$ the barber] that [$_{ph2}$ the banker] *praised* …) and proper nouns matching condition ([$_{ph1}$ John] that [$_{ph2}$ Dan] *praised* …) ranking higher in terms of difficulty (greatest slow down at verb segment), while pronouns ([$_{ph1}$ you] that [$_{ph2}$ we] *praised* …) are easier (fastest reading time). No CFG-based parsing algorithm (in fact, no classic algorithm implements the non-local dependencies in (2) as presented in §2) or Minimalist deductive parsing (parsing strategies exploit the weak equivalence of MGs with multiple Context Free Grammars, Michaelis 1998) have a chance to compare these cases.

## 3 A processing-friendly proposal

Phase-based Minimalist Grammars (PMGs, Chesi 2012) suitable for parsing of sentences like the ones in (2) can be formalized as follows:

(3) PMG able to parse cleft sentences

| *Lexicon* |
| --- |
| [[+D +Sg John$_i$] [N _*i*]], [[+D +Sg Dan$_i$] [N _*i*]], [N +Sg banker], [N +Sg lawyer], [+D the], [+D +P1 +Pl +case_acc me [N Ø]], [+D +P2 +Sg +case_nom you [N Ø]], [+T will], [+T that], [=[DP (+case_nom)] =[DP (+case_acc)] V meet], [+exp it], [=rCP BE is] |
| *Phases* |
| *DP* → [$_{DP}$ ([+F Ø]/[+S Ø]) +D N] |
| *Cleft* → [$_{CP}$ +Exp BE] |
| *rCP* → [$_{CP}$ +F +FIN (+S) +T V] |
| *Operations* |
| *Merge* = ([$_{phH}$ +f (+f$_n$) (H)], [+f L]) = [$_{phH}$ [+f L (+f$_n$) (H)]] |
| *Phase Projection* = [$_{phH}$ =**phX** H] = [$_{phH}$ ~~=**phX**~~ H [$_{\boldsymbol{phX}}$ ]] |
| *Move* = if expected [$_{phX}$ +f X] and found [$_{phX}$ [$_{phY}$ +f $\underline{+g}$ Y] X] → MEM([$_{phY}$ +g <Y>)]) |

As in MGs (Stabler 1997), the *Lexicon* is a finite set of lexical items storing phonetic, semantic (here ignored) and syntactic features (functional *+F*, selectional *=S*, categorial *C*); an item bearing a selection feature, e.g. [=XP A], requires an XP ph(r)ase right afterward: [~~=XP~~ A [XP ]] (once features are projected in the structure, i.e. [XP ], the selection features are deleted, i.e. ~~=XP~~); functional features, e.g. +X express a functional specification like determiner +D, tense +T or topic +S (when placed under brackets, e.g. (+f), functional features are optional; ø indicates phonetically null items).
*Merge* simply unifies the expected structure built so far with a new incoming item, if and only if, this item bears (at least) the first relevant feature expected (Merge operation is greedy: an item bearing more features in the correct expected order will lexicalize them all):

1. *Merge*([+X +Y +Z W ], [+X +Y A])=[[+X +Y A] +Z W ]
2. *Merge*([[+X +Y A] +Z W ], [+Z B])=[[+X +Y A][+Z B] W ]
3. *Merge*([[+X +Y A][+Z B] W ], [W C])=[[+X +Y A][+Z B] [W C]]

*Move* uses a Last-In-First-Out (LIFO) memory buffer (M) to create non-local dependencies: M is used to store unexpected bundles of features merged in the derivation (below, underlined features, e.g. [$\underline{+W\ U}$], are the unexcepted ones triggering Move):

1'. *Merge*([+X +Y +Z W ], [+X +W U A]) = [[+X $\underline{+W\ U}$ A] +Z W ]
2'. *Move*([+X $\underline{+W\ U}$ A]) = *M*[+W U <A>]

Items in the memory buffer M will be re-merged in the structure, before any other item taken from the lexicon, as soon as a coherent selection is introduced by another merged item:

3'. *Merge*([ … [W ~~=[+W U]~~ C [+W U ]]], *M*[+W U <A>]) =
[ … [W ~~=[+W U]~~ C [+W U <A>)]]], *M*[ *empty* ]

Notice that phonetic features (items under angled brackets, i.e. [<A>]) are not re-merged in the structure (that is, they are not expected to be found in the input) since they are already been pronounced/parsed in the higher position. When the M(emory) buffer is empty and no more selection features must be expanded, the procedure ends.

### 3.1 Parsing cleft structures with PMGs

The parsing algorithm using the minimalist grammar described in (3) implements an Earley-like procedure composed of three sub-routines:

1. *Ph(ase)P(rojection)* (Earley *Prediction* procedure): the most prominent (i.e. first/left most) select feature is expanded (the sentence parsing starts with a default PhP using one of the *phases* in grammar (3));
2. *Merge* (Earley *Scanning* procedure): if Memory is empty, the first available feature *F* in the expected phase is searched in the input/lexicon and possible items will be retrieved[2] (*search*(F) = [F $lex_1$], [F $lex_2$] … [F $lex_n$]) then unified with the expected structure (e.g. *Merge*([F ... X], [F $lex_1$]) = [[F $lex_1$]... X]); items stored in Memory are checked before the sentence input for Merge;
3. *Move*: if more features than the one expected are introduced, those features are clustered and moved in the LIFO Memory buffer: M[[*slot 1*][*slot 2*] … [*slot n*]].

Given the recursive, cyclic, application of the three subroutines above, this is the sequence of steps needed for parsing a cleft sentence like (2):

1. Default PhP (in this case: *Cleft*): [*CP* +Exp BE]
2. Search(+Exp): *M*[ *empty* ], Lex[[+exp it]]
3. Merge([*CP* +Exp BE], [+exp it]) = [*CP* [+exp it] BE]
4. Search(BE): *M*[ *empty* ], Lex[[BE is]]
5. Merge([*CP* [+exp it] BE], [=rCP BE is]) =
   [*CP* [+exp it] [=rCP BE is]]
6. PhP([*CP* [+exp it] [**=rCP** BE is]) =
   [*CP* [+exp it] [~~=rCP~~ BE is [*CP* +F +FIN +S +T V]]
7. Search(+F): *M*[ *empty* ], Lex[[*DP* [+F Ø] +D N]]
8. Merge([…[*CP* +F +FIN +S +T V]], [*DP* [+F Ø] +D N]) =
   [*CP* [*DP* [+F Ø] +D N] +FIN +S +T V]]
9. Search(+D): *M*[ *empty* ], Lex[[+D the]]
10. Merge([*DP* [+F Ø] +D N], [+D the]) =
    [*CP* [*DP* [+F Ø] [+D the] N] +FIN +S +T V]]
11. Search(N): *M*[ *empty* ], Lex[[N banker]]
12. Merge([*DP* [+F Ø] [+D the] N], [N banker]) =
    [*CP* [*DP* [+F Ø] [+D the] [N banker]] +FIN +S +T V]]
13. Move([*DP* [+F Ø] [+D the] N], [N banker]) =
    M[[*DP* +D N <the banker>]]
    (Move is triggered because at step 8 +D N were unexpected; only after full lexicalization [*DP* [+F Ø] +D N] is stored in M, namely at step 13)
14. Search(+FIN): *M*[[*DP* +D N <the banker>]], Lex[[+FIN that]]
15. Merge([*CP* [*DP* [+F Ø] [+D the] [N banker]] +FIN +S +T V]], [+FIN that]) = [*CP* [*DP* [+F Ø] [+D the] [N banker]] [+FIN that] +S +T V]]
16. Search(+S): *M*[[*DP* +D N <the banker>]], Lex[[*DP* [+S Ø] +D N]]
17. Merge([*CP* [*DP* [+F Ø] [+D the] [N banker]] [+FIN that] +S +T V]], [*DP* [+S Ø] +D N]) = [*CP* [*DP* [+F Ø] [+D the] [N banker]] [+FIN that] [*DP* [+S Ø] +D N] +T V]]
18. (repeat 9-13 mutatis mutandis)
19. Search(+T): *M*[[*DP* +D N (the lawyer)],[*DP* +D N (the banker)]], Lex[+T will]
20. Merge([*CP* [*DP* [+F Ø] [+D the] [N banker]] [+FIN that] [DP [+S Ø] [+D the] [N lawyer]] +T V]], [+T will]) = ([*CP* [*DP* [+F Ø] [+D the] [N banker]] [+FIN that] [DP [+S Ø] [+D the] [N lawyer]] [+T will] V]]
21. Search(V): *M*[[*DP* +D N (the lawyer)],[*DP* +D N (the banker)]], Lex[=DP =DP V meet]
22. Merge([*CP* [*DP* [+F Ø] [+D the] [N banker]] [+FIN that] [*DP* [+S Ø] [+D the] [N lawyer]] [+T will] V]], [=DP =DP V meet]) =
    [*CP* [*DP* [+F Ø] [+D the] [N banker]] [+FIN that] [*DP* [+S Ø] [+D the] [N lawyer]] [+T will] [=DP =DP V meet]]
23. PhP([*CP* … [**=DP** =DP V meet]]) = [*CP* … [~~=DP~~ =DP V meet [DP +D N]]]
24. Merge ([*CP* … [~~=DP~~ =DP V meet [*DP* +D N]]], M[[*DP* +D N (the lawyer)]] = ([*CP* … [~~=DP~~ =DP V meet [*DP* +D N (the lawyer)]]]
25. PhP([*CP* … [~~=DP~~ **=DP** V meet]]) = [*CP* … [~~=DP~~ ~~=DP~~ V meet [DP +D N (the lawyer)] [*DP* +D N]]]
26. Merge ([*CP* … [~~=DP~~ ~~=DP~~ V meet … [DP +D N]]], M[[*DP* +D N (the banker)]] = ([*CP* … [~~=DP~~ ~~=DP~~ V meet … [*DP* +D N (the banker)]]]

According to the lexicon and the phase expectations, step 10 and 19 could have found in the input [+D N John], [+D N Dan], [+D +P1 +Pl +case_acc me [N Ø]] or [+D +P2 +Sg +case_nom you [N Ø]], capturing all possible combinations of definite descriptions, correct pronominal DPs and proper nouns. Exactly all the possibilities we want to test.

## 4 Explaining the "similarity" problem in terms of cue-based feature retrieval

According to Warren & Gibson (2005) revealed reading times (see also Gordon et al. 2004 for very similar results) we can roughly rank on a difficulty scale all the (3x3) tested conditions (D = definite condition, e.g. "the banker", N = nominal condition, e.g. "Dan", P = pronoun condition, e.g. "we"; for instance D-D stands for "it is *the banker* that *the lawyer* will meet…", vs D-P condition "it is *the banker* that *we* will meet…"):

(4) $$\text{D-D} \geq \text{N-D} \approx \text{N-N} \approx \text{P-D} > \text{D-N} \geq \text{P-N} > \text{D-P} \geq \text{N-P} \approx \text{P-P}$$

Building on Gillund & Shiffrin (1984) Search of Associative Memory (SAM) model, and assuming a cue-based retrieval mechanism for items in memory (Van Dyke & McElree 2006), we can define a complexity (*C*) function associated to the features to be retrieved from M (Feature Retrieval

---

[2] For reason of space, I will not discuss here neither lexical and syntactic ambiguity nor reanalysis (i.e. recovery from wrong expectations); the proposed algorithm here is meant to be a Top-Down complete procedure, that is, all the possible ambiguities will be taken into consideration and stored in the parsing "chart" as in the classic Earley's parser. For ranking of alternatives see Hale (2001).

Cost, *FRC*, Chesi 2016) for each item to be re-merged after the phase projection at verb (*V*):

(5) $C_{FRC}(V) = \prod_{i=1}^{m_x} \frac{(1+nF_i)^{m_i}}{(1+dF_i)}$

In the formula above, *m* is number of items stored in memory at retrieval, *nF* is the number of features characterizing the argument to be retrieved that are non-distinct in memory (i.e. also present in other objects in memory), *dF* is number of distinct cued features (e.g. case features explicitly probed by the verb selection). $C_{FRC}$ will express the cost, in numerical terms, that should fit with the revealed reading time (i.e. higher differences in reading times, higher differences in $C_{FRC}$).
According to the lexicon in (3), the cost for retrieving the correct items in the D-D condition, for instance, is calculated as follows:

1. [=[DP (+case_nom)] =DP(+case_acc) V meet] will trigger retrieval of the first item (the last inserted one in the buffer) which is (step 24) the *DP* [+D +Sg N (the lawyer)]
2. No cued-features are present (the verb selection only asks for an optional nominative case) and the 3 features to be retrieved are in fact shared with the other item in memory ([+D +Sg N (the banker)])
3. Hence: $C_{FRC} = \frac{(1+3)^2}{(1+0)} \text{ x } \frac{(1+0)^1}{(1+0)} = 16$

Notice that retrieving the object when the subject has been removed from memory has a minimal cost since no confounding features are present anymore in memory. As for the other relevant conditions: N-N, as in D-D condition share the same features hence we expect them to have similar cost except for the fact that N feature is not fully lexicalized, but it is a trace of an N-to-D movement (Longobardi 1994). Counting this as 0.5 (further investigation is needed to correctly assign a cost to an emptied lexical position), we obtain 12,25. Same complexity for N-D condition (since the [N] lexical feature in D is compared to the trace [N _*i*] feature of N counting 0.5). While we would expect slightly smaller cost with the P-D condition (P does have a [N ø] empty feature), that is 9, we will correctly predict simpler complexity for retrieving pronouns at the subject position, since they are always bearing person features (which are distinct from default 3rd person of D and N) and they are marked for case (which is cued by the verb, producing the minimal cost in the P-P condition ($C_{FRC}$= 1) and similar costs in the D-P and N-P conditions (both $C_{FRC}$=4). Predictions can be further differentiated by adding a cost for encoding the features in the structure (*eF*) which is (to keep the calculation as simple as possible) proportional to the number of lexical features to be encoded once an item is retrieved from memory (the numerator of the $C_{FRC}$ cost function becomes: $(1 + nF_i + eF_i)^{m_i}$). This corresponds to an increase of +1 for D and +0,5 for N at retrieval. The new $C_{FREC}(V)$ in the different conditions becomes:

$$C_{FREC}(V)_{\text{D-D}} = \frac{(1+3+1)^2}{(1+0)} \text{ x } \frac{(1+0+1)^1}{(1+0)} = 50$$
$$C_{FREC}(V)_{\text{N-D}} = \frac{(1+3+1)^2}{(1+0)} \text{ x } \frac{(1+0+0,5)^1}{(1+0)} = 37,5$$
$$C_{FREC}(V)_{\text{N-N}} = \frac{(1+3+0,5)^2}{(1+0)} \text{ x } \frac{(1+0+0,5)^1}{(1+0)} = 30,37$$
$$C_{FREC}(V)_{\text{P-D}} = \frac{(1+3+1)^2}{(1+0)} \text{ x } \frac{(1+0+0)^1}{(1+0)} = 25$$
$$C_{FREC}(V)_{\text{D-N}} = \frac{(1+2+0,5)^2}{(1+0)} \text{ x } \frac{(1+0+1)^1}{(1+0)} = 24,5$$
$$C_{FREC}(V)_{\text{P-N}} = \frac{(1+2+0,5)^2}{(1+0)} \text{ x } \frac{(1+0+0)^1}{(1+0)} = 12,25$$
$$C_{FREC}(V)_{\text{D-P}} = \frac{(1+1+0)^2}{(1+1)} \text{ x } \frac{(1+0+1)^1}{(1+0)} = 8$$
$$C_{FREC}(V)_{\text{N-P}} = \frac{(1+1+0)^2}{(1+1)} \text{ x } \frac{(1+0+0,5)^1}{(1+0)} = 6$$
$$C_{FREC}(V)_{\text{P-P}} = \frac{(1+1+0)^2}{(1+1)} \text{ x } \frac{(1+0+0)^1}{(1+0)} = 2$$

Though in some cases *FREC* predicts slightly larger differences (e.g. *D-D* vs *N-D/N-N* condition), it correctly ranks all conditions revealed by the discussed experiment, and it is coherent with specific predictions (e.g. related to feature matching) discussed in literature (Belletti & Rizzi 2013).

## 5 Conclusion

In this paper I presented an adaptation of Earley's Top-Down parsing algorithm to be used with a simple implementation of a Minimalist Grammar (PMG). The advantages of this approach are both in terms of cognitive plausibility and parsing/performance transparency. From the cognitive plausibility perspective, I showed how a re-orientation of the minimalist structure building operations Merge and Move is sufficient to include such operations directly within a parsing procedure. This is a step toward the "Parser Is the Grammar" (PIG) default hypothesis (Phillips 2006) and a welcome simplification of the linguistic competence description: such a grammar description (i.e. our linguistic competence) is shared both in production (generation) and in comprehension (parsing); this seems trivial from a cognitive perspective (we have a unique Broca's area activated in syntactic processing both in parsing and in generation), but it is far from trivial in computational terms. On the other hand, from the parsing/performance transparency perspective, I presented a complexity metric (FREC), based on cued features stored in memory which better characterize performance in object clefts constructions compared to alternative models: for instance the Depencency Locality Theory (DLT) based on accessibility hierarchy (Gibson 2000) is unable to predict high complexity in N-N condition comparable to N-D or D-D condition, since N should be uniformly more accessible than D, contrary to the facts. The proposed model, obviously should be extended in many respects to capture other critical phenomena (see Lewis & Vasishth 2005) but the first results on specific well-studied constructions, like object clefts, seem very promising.